\documentclass[sigconf,nonacm]{acmart}

\setcopyright{none}
\acmYear{2027}
\copyrightyear{2027}
\acmConference[KDD '27]{The 33rd ACM SIGKDD Conference on Knowledge Discovery and Data Mining}{August 1--5, 2027}{San Jose, CA, USA}

\usepackage{booktabs}
\usepackage{multirow}
\usepackage{amsmath}
\usepackage{enumitem}
\usepackage{float}
\usepackage{stfloats}
\usepackage{placeins}
\usepackage{xcolor}

\newcommand{\bench}{ECG-InterpBench}
\newif\ifresultsaudited

\providecommand{\MSCells}{450}
\providecommand{\MSMatchedBlocks}{75}

\providecommand{\MSReconTopModel}{HuBERT-ECG}
\providecommand{\MSReconOrder}{HuBERT-ECG $>$ ECG-JEPA $>$ CSFM $>$ ECG-FM $>$ CARDIAC-FM $>$ ST-MEM}
\providecommand{\MSReconSignificantPairs}{15}
\providecommand{\MSReconMedianScaleTau}{1.00}
\providecommand{\MSReconMinScaleTau}{1.00}

\providecommand{\MSReconTopAUC}{0.985}
\providecommand{\MSReconTopCILow}{0.985}
\providecommand{\MSReconTopCIHigh}{0.985}
\providecommand{\MSReconTopRankOne}{1.000}
\providecommand{\MSSemanticTopModel}{ECG-JEPA}
\providecommand{\MSSemanticOrder}{ECG-JEPA $>$ ECG-FM $>$ CARDIAC-FM $>$ ST-MEM $>$ HuBERT-ECG $>$ CSFM}
\providecommand{\MSSemanticSignificantPairs}{13}
\providecommand{\MSSemanticMedianScaleTau}{0.73}
\providecommand{\MSSemanticMinScaleTau}{0.73}

\providecommand{\MSCoverageOrder}{CARDIAC-FM $>$ ECG-FM $>$ ECG-JEPA $>$ ST-MEM $>$ HuBERT-ECG $>$ CSFM}
\providecommand{\MSCoverageSignificantPairs}{13}
\providecommand{\MSCoverageMedianScaleTau}{0.73}
\providecommand{\MSCoverageMinScaleTau}{0.60}

\providecommand{\MSStabilityTopModel}{ECG-FM}
\providecommand{\MSSubspaceTopModel}{CSFM}

\providecommand{\MSStabilityTopAUC}{0.371}
\providecommand{\MSSubspaceTopAUC}{0.757}

}

\ifresultsaudited\else
  \renewcommand{\includegraphics}[2][]{\fbox{\parbox[c][2.8cm][c]{0.82\linewidth}{\centering Audited result figure pending}}}
\fi

\begin{document}

\title[ECG-InterpBench]{ECG-InterpBench: Benchmarking the Interpretability of ECG Foundation Models with Matched-Scale Sparse Autoencoders}

\author{Yixuan Duan}
\affiliation{%
  \department{Department of Electrical and Computer Engineering}
  \institution{Rice University}
  \city{Houston}
  \state{Texas}
  \country{USA}}
\email{yd68@rice.edu}

\author{Wei Qiu}
\authornote{Corresponding author.}
\affiliation{%
  \department{Department of Electrical and Computer Engineering}
  \institution{Rice University}
  \city{Houston}
  \state{Texas}
  \country{USA}}
\email{wq8@rice.edu}

\renewcommand{\shortauthors}{Duan and Qiu}

\begin{abstract}
Existing benchmarks for electrocardiogram foundation models primarily evaluate downstream predictive performance, providing limited insight into whether their internal representations can be faithfully decomposed, clinically interpreted, or reproduced across independent analyses. We introduce \bench, a benchmark designed to systematically evaluate the interpretability of ECG foundation-model representations. \bench{} uses sparse autoencoders as standardized measurement instruments and matches their capacity across models to enable controlled comparisons. We evaluate six frozen ECG foundation models across five standardized encoder depths, five matched dictionary widths, and three random seeds, producing a \MSCells{}-cell interpretability atlas comprising \MSMatchedBlocks{} exactly matched six-model comparison blocks.  The benchmark evaluates complementary dimensions of representation interpretability, including sparse reconstruction fidelity, single-feature accessibility and coverage of 49 clinically meaningful ECG measurements, and cross-seed feature reproducibility. The evaluation further quantifies patient-sampling uncertainty, depth- and seed-dependent variation, and sensitivity to the sparsity parameterization. The benchmark reveals that ECG foundation models exhibit distinct interpretability profiles. A matched replication on MIMIC-IV-ECG confirms that reconstruction fidelity and clinical accessibility identify different leading models. The benchmark is accompanied by executable evaluation code, standardized manifests, cell-level metrics, and reproducibility audits. \bench{} complements performance-centered ECG benchmarks by providing a capacity-controlled and reproducible framework for comparing ECG foundation models across distinct dimensions of representation interpretability.
\end{abstract}



\maketitle

\section{Introduction}

\begin{figure*}[!t]
  \centering
  \includegraphics[width=\textwidth]{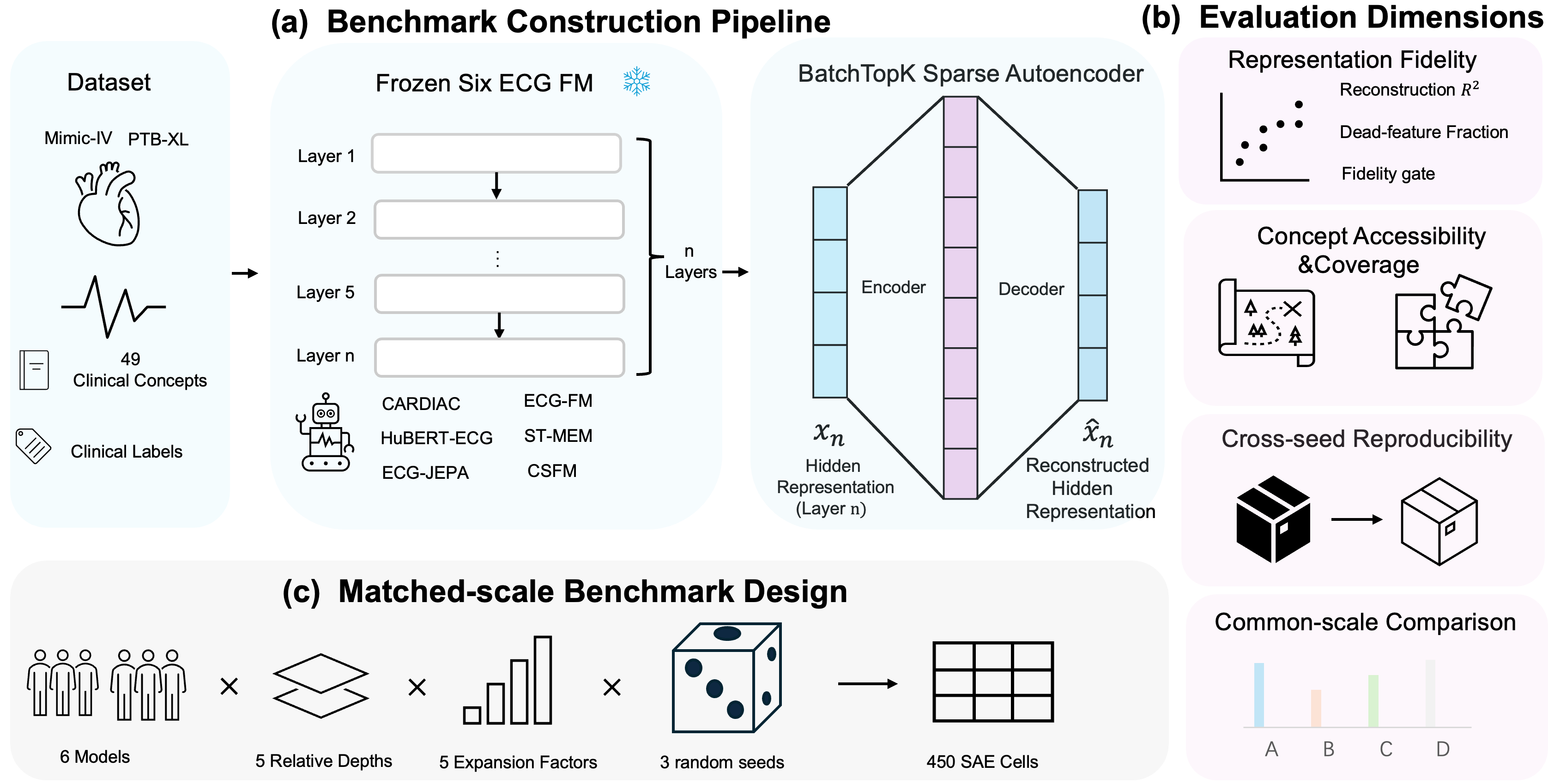}
  \caption{Overview of \bench. (a) PTB-XL and credentialed MIMIC-IV ECGs are passed through six frozen ECG FMs, and layer representations are decomposed with BatchTopK SAEs. (b) Each representation is evaluated separately for reconstruction fidelity, single-feature clinical concept accessibility and coverage, and cross-seed reproducibility. (c) The matched-scale design crosses six models, five relative depths, five expansion factors, and three seeds, producing 450 SAE cells for common-scale model comparison.}
  \Description{The benchmark pipeline has three parts. First, ECG datasets and clinical concepts are processed by six frozen ECG foundation models, and hidden representations from multiple layers are reconstructed by BatchTopK sparse autoencoders. Second, the resulting representations are evaluated for reconstruction fidelity, concept accessibility and coverage, and cross-seed reproducibility. Third, six models, five relative depths, five expansion factors, and three random seeds form a matched 450-cell comparison grid.}
  \label{fig:multiscale-workflow}
\end{figure*}

ECG foundation models (ECG FMs) learn general-purpose representations that support diverse rhythm, morphology, and cardiovascular risk-prediction tasks~\cite{mckeen2025ecgfm,gu2026csfm,na2024stmem}. Existing evaluations, however, focus primarily on downstream accuracy, transfer performance, or label efficiency~\cite{mehari2022ssl,kiyasseh2021clocs,vaid2023heartbeit}. These outcomes do not reveal how the hidden representation is organized: whether it can be faithfully decomposed, whether clinically relevant measurements are localized to identifiable features, or whether the same structure is recovered across independent decompositions. This gap matters in clinical settings because a shared representation can propagate acquisition artifacts, demographic confounding, or cohort-specific shortcuts to every downstream model built on it~\cite{rudin2019blackbox,ghassemi2021falsehope,tonekaboni2019clinicians}.

Existing interpretability tools address related but different questions. Output attribution identifies waveform regions that influence a prediction, but it does not audit the organization of a task-general hidden representation and may fail parameter-randomization or other sanity checks~\cite{strodthoff2019mi,adebayo2018sanity}. Linear probes establish whether a clinical variable is decodable, but not whether it is localized to a small number of features, preserved by a faithful sparse decomposition, or recovered across independent SAE trainings. Representation-level interpretability is therefore not equivalent to predictive accuracy, linear decodability, or visually plausible saliency; it requires explicit, complementary measurements~\cite{murdoch2019definitions}.

Sparse autoencoders (SAEs) provide a scalable instrument for this audit by mapping dense activations into sparse codes and reconstructing the original representation from learned decoder directions~\cite{cunningham2024sae,gao2025scaling,bussmann2024batchtopk}. Their behavior depends on dictionary width and sparsity, making capacity matching central to controlled cross-model measurement~\cite{templeton2024scaling,lieberum2024gemmascope,makelov2025principled}. \bench{} therefore evaluates every FM with the same family of dictionary scales and sparsity settings and summarizes all models over common scale support. This design separates differences among FM representations from differences in SAE measurement capacity.

\bench{} uses a matched family of SAEs as the standardized measurement instrument and treats the frozen FM representation as the object being evaluated. We operationalize representation interpretability along three separately reported axes: (1) sparse reconstruction fidelity and dictionary utilization; (2) train-selected single-feature accessibility and coverage of clinical ECG concepts; and (3) cross-seed reproducibility of decoder features and selected feature subspaces. Together, these axes provide a structured profile of how faithfully, locally, and reproducibly each representation can be interpreted.

We evaluate six frozen ECG FMs at five standardized relative depths, five matched SAE scales, and three random seeds. The resulting 450-cell atlas contains 75 comparison blocks in which all six models share the same target relative depth, dictionary capacity, sparsity protocol, and seed. Comparisons are made only within these blocks and summarized over the same scale support for every model. We additionally quantify patient-sampling and design uncertainty and test sensitivity to an alternative sparsity parameterization. Figure~\ref{fig:multiscale-workflow} summarizes this capacity-controlled design.

The benchmark reveals distinct, metric-specific interpretability profiles. Reconstruction fidelity, single-feature clinical accessibility, and decoder reproducibility expose complementary strengths across ECG FMs. A matched MIMIC-IV-ECG replication preserves the separation between reconstruction and accessibility leaders. The resulting profile supports direct model comparison across all measured representation properties.

The benchmark implementation, experiment protocols, and manuscript artifacts are publicly available in our \href{https://github.com/JayDuan123/2027-kdd-ECG-InterpBench}{GitHub repository}.

Our contributions are:
\begin{itemize}[leftmargin=1.3em,itemsep=1pt,topsep=2pt]
  \item We introduce a capacity-controlled benchmark for comparing representation-level interpretability across six ECG FMs under a matched SAE protocol.
  \item We construct a 450-cell matched-scale interpretability atlas that maps how representation interpretability changes across encoder depth and SAE scale.
  \item We define leakage-controlled interpretability metrics spanning sparse fidelity, single-feature clinical accessibility and coverage, and cross-seed feature and subspace reproducibility.
  \item We provide robust interpretability comparisons supported by patient and design uncertainty, sparsity sensitivity, and external-cohort evaluation.
\end{itemize}

\section{Related Work}

\paragraph{ECG foundation models.}
Supervised deep ECG systems established high diagnostic performance for arrhythmia, multi-label 12-lead interpretation, and ventricular-dysfunction screening~\cite{hannun2019cardiologist,ribeiro2020automatic,attia2019screening}. PTB-XL subsequently enabled reproducible architecture benchmarking and transfer analysis~\cite{strodthoff2021benchmark}, while contrastive, predictive, and masked-reconstruction objectives showed that useful ECG representations can be learned without task labels~\cite{mehari2022ssl,kiyasseh2021clocs,vaid2023heartbeit}.

The evaluated encoders use multimodal, contrastive, predictive, clustering, or spatio-temporal objectives. They are CSFM, CARDIAC-FM, ECG-FM, ECG-JEPA, HuBERT-ECG, and ST-MEM~\cite{gu2026csfm,li2026cardiacfm,mckeen2025ecgfm,kim2024ecgjepa,coppola2024hubert,na2024stmem}. Their released channel selection, sample rate, duration, tokenization, and pooling APIs differ. We preserve those interfaces and standardize only the comparison coordinates---relative depth and SAE capacity.

\paragraph{Sparse representation analysis.}
SAEs are motivated by sparse coding~\cite{olshausen1997sparse,mairal2009dictionary} and have been used to recover interpretable directions from language, biological, and other foundation models~\cite{bricken2023monosemanticity,simon2025interplm}. Large-scale dictionary releases and scaling studies show that dictionary width, sparsity, reconstruction, and feature quality interact~\cite{cunningham2024sae,gao2025scaling,templeton2024scaling,lieberum2024gemmascope}; BatchTopK makes the batch-level activation budget explicit~\cite{bussmann2024batchtopk}. Evaluations of SAE fidelity, sparse probing, and intervention further caution that no single metric establishes interpretability~\cite{makelov2025principled,karvonen2025saebench}. These results motivate a capacity sweep, but they also imply that a cross-model study must match the sweep rather than select a separate favorable capacity for each encoder.

Prior work applies TopK SAEs to EEG transformers across layers and scales, evaluating dictionary health, clinical semantics, steering, and spectral decoding~\cite{lehnschioler2026eegsae}. \bench{} focuses instead on controlled interpretability comparisons across ECG foundation models. It restricts primary comparisons to common scales, integrates identical scale support for every model, pairs patient resamples across models, freezes concept selection before test evaluation, and evaluates seed-level feature identity against random matching.

\paragraph{Concept and benchmark evaluation.}
Interpretability depends on the intended audience and use, and medical explanations require evaluation against clinical needs rather than visual plausibility alone~\cite{murdoch2019definitions,rudin2019blackbox,tonekaboni2019clinicians,ghassemi2021falsehope}. ECG attribution studies can recover diagnostically relevant waveform regions~\cite{strodthoff2019mi}, but output attribution answers a different question from auditing the organization of a task-general hidden representation. Concept activation vectors define supervised directions~\cite{kim2018tcav}, concept bottlenecks expose human-specified intermediate variables~\cite{koh2020concept}, and LEACE removes linearly decodable subspaces in closed form~\cite{belrose2023leace}. These methods likewise answer different questions from an unsupervised sparse dictionary.

Explanation methods can pass qualitative inspection while failing sanity, leakage, or control tests~\cite{adebayo2018sanity,jethani2023leakage}. SAE studies accordingly report heterogeneous criteria, including one-dimensional binary probe loss, automated explanation, TCAV, intervention selectivity, and reconstruction fidelity~\cite{gao2025scaling,lehnschioler2026eegsae,karvonen2025saebench}. Those values are not numerically commensurate with held-out Pearson correlation for continuous ECG measurements, so we do not assign our correlation scale a literature-wide ``strong alignment'' threshold. Our benchmark instead uses it as a leakage-controlled relative comparison under matched ECG-FM depth--scale blocks, while retaining every failed cell in a fixed denominator.

\section{Benchmark Design}
\label{sec:benchmark-design}

\subsection{Source cohorts, frozen models, and representations}

PTB-XL supplies 21,799 twelve-lead ECGs and 49 waveform-derived measurement and morphology concepts~\cite{wagner2020ptbxl}. We use the existing patient-disjoint split: 15,149 records for training, 3,325 for validation, and 3,325 for test. The test set contains 2,862 unique patients. Every model receives the same ordered record manifest, and preprocessing statistics are estimated on training records only. PTB-XL is the source cohort for the primary benchmark. For external validation, we repeat the complete 450-cell multiscale SAE audit on a credentialed, patient-partitioned MIMIC-IV-ECG manifest of 100,000 ECGs from 48,491 patients (Appendix~\ref{app:mimic-replication}).

All six pretrained encoders remain frozen throughout the benchmark. We update no FM parameters and perform no task-specific fine-tuning; only the SAEs are trained on activations extracted from the fixed checkpoints. Each encoder is called through its released preprocessing path (Appendix~\ref{app:model-interfaces}, Table~\ref{tab:multiscale-models}). We audit pre-projection transformer-layer states for CARDIAC-FM; its released 768-to-512 pooled projection is used only in the downstream stage and is not the dimension of the source layer atlas. Consequently, all six source layer atlases have $d=768$.

\begin{figure*}[!t]
  \centering
  \includegraphics[width=\textwidth]{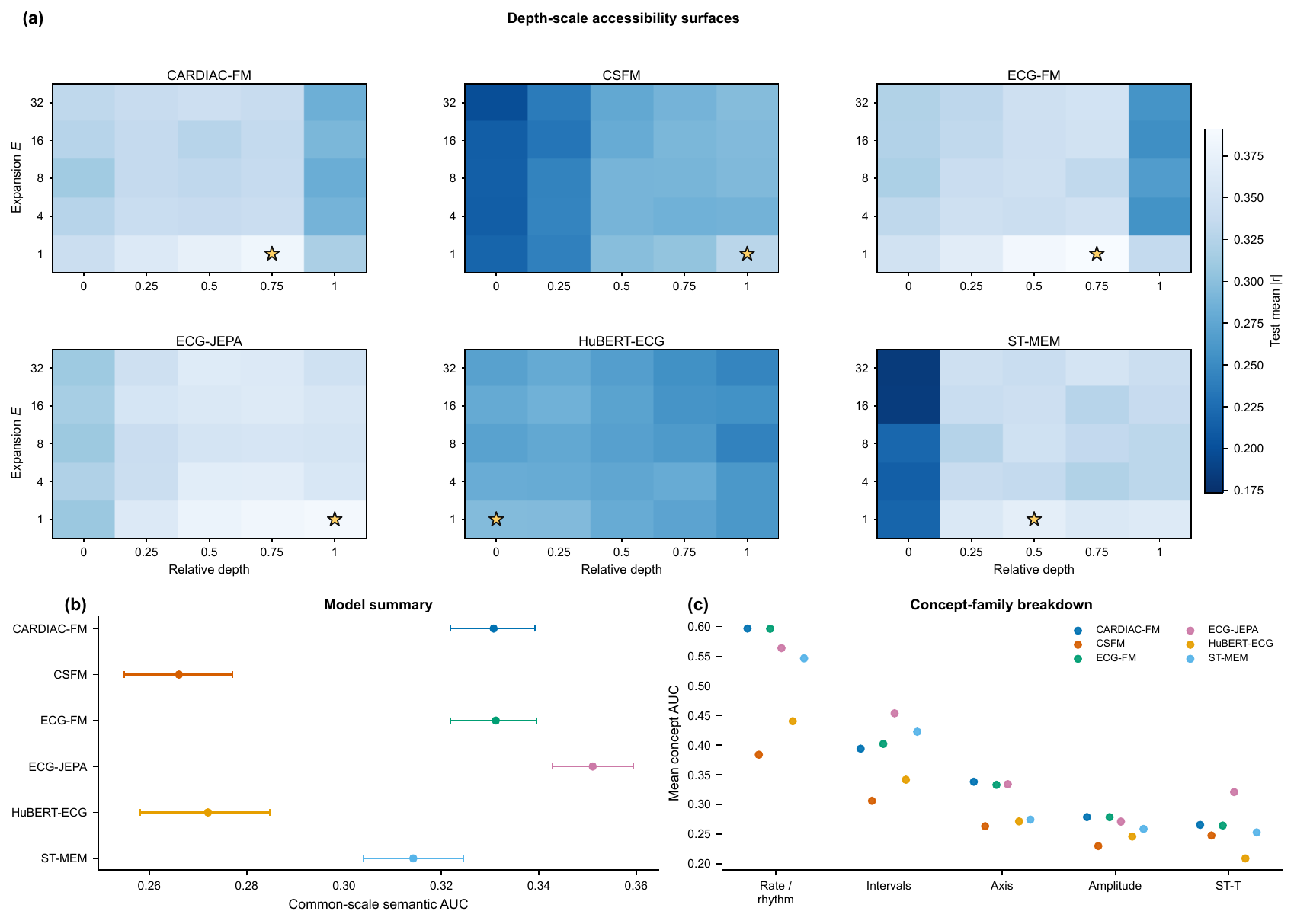}
  \caption{Single-coordinate clinical accessibility across the matched SAE atlas. (a) Test mean absolute concept correlation over five relative depths and five exactly matched expansion factors; circles mark each model's peak cell. (b) Common-scale semantic AUC with paired patient-cluster 95\% intervals. (c) Mean concept-level common-scale AUC by ECG concept family. Features are selected on the training subset and frozen before held-out evaluation.}
  \Description{A three-panel figure shows six depth-scale heatmaps for ECG foundation models, a forest plot of common-scale semantic AUC with confidence intervals, and a dot plot of mean concept accessibility by ECG concept family.}
  \label{fig:semantic-atlas}
\end{figure*}

\begin{figure*}[!t]
  \centering
  \includegraphics[width=\textwidth]{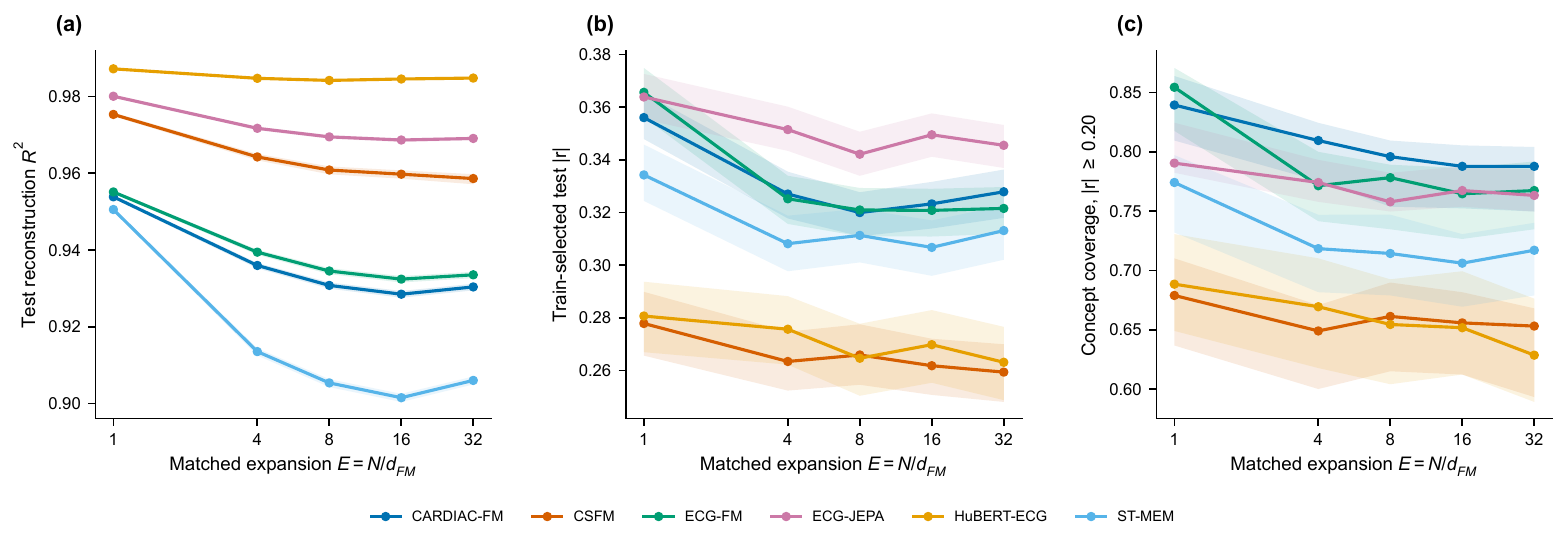}
  \caption{Patient-bootstrap FM profiles at each common SAE scale: (a) reconstruction $R^2$, (b) train-selected single-coordinate clinical accessibility, and (c) concept coverage at $|r|\geq0.20$. Lines compare models at identical $E$ and therefore identical absolute $N$; bands are paired 95\% patient-cluster intervals.}
  \Description{Three line panels show reconstruction, train-selected single-coordinate clinical accessibility, and concept coverage for six models over five matched expansion factors, with confidence bands.}
  \label{fig:matched-curves}
\end{figure*}

\subsection{Standardized depth and exact scale matching}

For an encoder exposing $L$ audited states, target relative depth $q\in\{0,.25,.5,.75,1\}$ maps to
\begin{equation}
  \ell(q)=\left\lfloor q(L-1)+\tfrac{1}{2}\right\rfloor.
\end{equation}
We record both $q$ and the realized depth $\ell/(L-1)$. CSFM maps to layers $(0,1,3,4,5)$; each 12-state encoder maps to $(0,3,6,8,11)$. Relative depth provides a shared coordinate for aligning early, intermediate, and final representation stages across encoders with different layer counts. This alignment enables systematic depth-wise comparison of where reconstruction fidelity, clinical accessibility, and reproducible sparse structure emerge across models.

For a batch of standardized activations $X\in\mathbb{R}^{B\times d}$, BatchTopK computes positive pre-activations and retains the largest $Bk$ values over the complete batch:
\begin{align}
  P &= \operatorname{ReLU}\!\left((X-b_d)W_e^{\top}+b_e\right),\\
  Z &= \operatorname{BatchTopK}_{Bk}(P), \qquad \widehat{X}=ZW_d^{\top}+b_d.
\end{align}
Decoder columns are constrained to unit norm. Thus $k$ specifies the batch-average active-feature budget, while the reported realized mean $L_0$ summarizes per-record activation sparsity.

The primary grid fixes expansion $E=N/d\in\{1,4,8,16,32\}$ and $k/d=1/8$. Because $d=768$ for every source layer, all models use exactly the same widths
\begin{equation}
  N\in\{768,3072,6144,12288,24576\}
\end{equation}
and the same $k=96$ at every scale. Each SAE is trained for 8,000 steps with batch size 256, Adam learning rate $3\times10^{-4}$, and seed 4311, 4312, or 4313. The complete grid contains
\begin{equation}
  6\ \text{models}\times5\ \text{depths}\times5\ \text{scales}\times3\ \text{seeds}=450
\end{equation}
cells, arranged as 75 blocks in which all six FMs share the same target depth, $N$, $k$, and seed. Primary contrasts pair all models at the same prespecified $E$ and summarize performance over the shared scale grid, with scale choices fixed before test evaluation.

\subsection{Leakage-controlled single-coordinate clinical accessibility}

Concept values are aligned by ECG identifier and standardized with training-split means and standard deviations. A missing or non-finite value is replaced by the training mean, which is zero in standardized coordinates. Coordinate selection uses the same deterministic 4,096-record subset of the training split in every cell. For concept $c$ and SAE code coordinate $j$, let $r^{\mathrm{train}}_{jc}$ denote their correlation on that subset. We select
\begin{equation}
  j_c^*=\arg\max_j |r^{\mathrm{train}}_{jc}|
\end{equation}
once and evaluate the same coordinate on validation and test records. The test single-coordinate accessibility score is the mean $|r^{\mathrm{test}}_{j_c^*c}|$ over 49 concepts; coverage is the fraction with $|r^{\mathrm{test}}_{j_c^*c}|\geq .20$. Training-only selection fixes every SAE coordinate, layer, and scale before held-out evaluation. This estimand quantifies how strongly one sparse coordinate locally exposes each concept, and the $.20$ cutoff provides a consistent descriptive operating point across models.

\subsection{Fixed-scale accessibility calibration}

At the common $E=8$ cells ($N=6144,k=96$), we compare seven readouts: fixed-$\alpha=10$ ridge from the dense state, full SAE code, or 16/four train-ranked SAE coordinates; and fit-free train-selected native, SAE, or random coordinates. Dense coordinates are evaluated once per model--depth, whereas SAE metrics average three seeds. The random control averages 20 shared seeds, each projecting the normalized state onto $N$ Gaussian unit directions before the same ReLU, BatchTopK, batch order, and training-only selection. We report held-out mean $|r|$, coverage at $|r|\geq.20$, paired differences, and random-seed percentile intervals. Coordinates and readout settings are fixed from the training data before validation and test outcomes are used for held-out evaluation.

\subsection{Metric-specific FM profiles}

\paragraph{Measurement fidelity.}
We report normalized reconstruction $R^2=1-\mathrm{SSE}/\mathrm{SST}$, dead-feature fraction, realized mean $L_0$, and the fraction of layer--scale cells satisfying the preregistered validation fidelity gate ($R^2\geq.90$ and dead fraction $<.20$). Reconstruction quantifies how faithfully the sparse instrument preserves a representation and is interpreted alongside clinical accessibility.

\paragraph{Single-coordinate clinical accessibility.}
Train-selected test correlation and concept coverage quantify how locally the fixed concept panel is exposed by individual sparse-code coordinates. Their primary role is relative model comparison under an identical selection and capacity protocol. Joint interpretation with reconstruction fidelity connects local concept exposure to the quality of the underlying sparse decomposition.

\paragraph{Reproducibility.}
Within each model, depth, and scale, we retain up to the 256 most active non-dead decoder directions per seed according to validation firing rate. Hungarian matching maximizes signed decoder cosine for each seed pair. We report matched cosine above a 100-permutation random-pairing floor and normalized overlap between the selected decoder subspaces.

\paragraph{Common-scale summary.}
For metric curve $m(E)$, the model profile is the normalized log-scale area
\begin{equation}
  \operatorname{AUC}_{E}(m)=
  \frac{1}{\log 32-\log 1}\int_{\log 1}^{\log 32}m(e^u)\,du,
\end{equation}
Here, $E$ is the SAE expansion factor, evaluated at $\{1,4,8,16,32\}$, and $m(E)$ may denote reconstruction $R^2$, single-coordinate mean $|r|$, concept coverage, matched decoder cosine, or selected-subspace overlap. We approximate the integral by trapezoidal integration over these five points in $\log E$. Division by $\log 32-\log 1$ makes $\operatorname{AUC}_{E}$ a normalized average across SAE scales. The resulting scale summary is then averaged over the five depths and three seeds as specified by each inferential analysis. Fidelity, single-coordinate accessibility, coverage, and stability jointly define the model's multiscale interpretability profile.

\subsection{Paired uncertainty and multiplicity}

The primary test inference resamples PTB-XL patients with replacement 2,000 times and retains all records belonging to a sampled patient. The sorted patient set and bootstrap seed are identical across all 450 cells, enabling paired FM differences. BatchTopK codes are computed once in the frozen original test order with batch size 256; bootstrap weights are then applied to patient-aggregated sufficient statistics, preserving a common encoding across all resamples. Reconstruction uses the frozen full-test reference mean in the SST term for every bootstrap draw. These intervals quantify held-out patient sampling uncertainty conditional on the frozen reference, FM and SAE weights, and train-selected coordinates. The three SAE seeds are averaged within each draw.

A second 10,000-sample crossed bootstrap resamples the five depth units and three SAE seeds to quantify design variation after integrating the frozen scale curve. Pairwise model contrasts are adjusted by Benjamini--Hochberg within each metric family~\cite{benjamini1995fdr}. Scale sensitivity is summarized by Kendall rank correlation between every pair of common $E$ values. Together, patient and design intervals characterize complementary sources of uncertainty: held-out patient sampling and depth--seed design variation.

\subsection{Sparsity sensitivity}

The primary fixed-$k/d$ arm holds the absolute active budget constant while dictionary width grows. A preregistered mid-depth sensitivity arm instead fixes $k/N=1/64$, giving $k\in\{12,48,96,192,384\}$ over the same five exact widths, six models, and three seeds. At relative depth $q=0.5$, this arm retains the primary training protocol and uses $N=\{768,3072,6144,12288,24576\}$ with $k=N/64=\{12,48,96,192,384\}$. It evaluates the stability of metric-specific FM ordering under an alternative sparsity parameterization, with both arms and their comparison preregistered before test evaluation.

\begin{table*}[t]
  \centering
  \caption{Final-layer input-harmonization audit. $\Delta$ values are changes in test mean $|r|$ from the native protocol. Coverage is the joint-protocol fraction of 49 concepts reaching $|r|\geq0.20$; CKA compares joint and native test representations.}
  \label{tab:input-harmonization}
  \begin{tabular}{lrrrrrr}
    \toprule
    Model & Native mean $|r|$ & $\Delta$ Lead & $\Delta$ Temporal & $\Delta$ Joint & Joint coverage & Joint CKA \\
    \midrule
    CARDIAC-FM & 0.2922 & $+0.0000$ & $-0.0058$ & $-0.0058$ & 0.857 & 0.991 \\
    CSFM       & 0.3321 & $-0.0004$ & $+0.0000$ & $-0.0004$ & 0.837 & \textbf{1.000} \\
    ECG-FM     & 0.3166 & $-0.0000$ & $-0.0048$ & $-0.0048$ & \textbf{0.959} & 0.997 \\
    ECG-JEPA   & \textbf{0.4380} & $+0.0000$ & $+0.0000$ & $+0.0000$ & \textbf{0.959} & \textbf{1.000} \\
    HuBERT-ECG & 0.2923 & $\boldsymbol{+0.0006}$ & $\boldsymbol{+0.0005}$ & $\boldsymbol{+0.0006}$ & 0.735 & \textbf{1.000} \\
    ST-MEM     & 0.3891 & $-0.0000$ & $-0.0131$ & $-0.0131$ & 0.898 & 0.770 \\
    \bottomrule
  \end{tabular}
\end{table*}

\begin{table*}[t]
  \centering
  \small
  \caption{MIMIC-IV-ECG held-out multiscale SAE profiles. Reconstruction, semantic alignment, and coverage are normalized expansion-curve AUCs averaged over five relative depths. Stability is matched decoder-feature cosine across SAE seeds; subspace stability is normalized overlap of the selected decoder subspaces. Higher is better except for dead-feature fraction.}
  \label{tab:mimic-model-profiles}
  \begin{tabular}{lrrrrrr}
    \toprule
    Model & Recon. & Semantic & Coverage & Dead fraction & Feature stability & Subspace stability \\
    \midrule
    CARDIAC-FM & 0.958 & 0.295 & 0.674 & 0.166 & 0.366 & 0.718 \\
    CSFM       & 0.979 & 0.261 & 0.582 & 0.380 & 0.293 & \textbf{0.778} \\
    ECG-FM     & 0.955 & 0.282 & 0.625 & \textbf{0.099} & \textbf{0.378} & 0.702 \\
    ECG-JEPA   & 0.985 & \textbf{0.324} & \textbf{0.781} & 0.447 & 0.340 & 0.748 \\
    HuBERT-ECG & \textbf{0.990} & 0.279 & 0.610 & 0.546 & 0.323 & 0.762 \\
    ST-MEM     & 0.960 & 0.274 & 0.600 & 0.211 & 0.348 & 0.760 \\
    \bottomrule
  \end{tabular}
\end{table*}

\section{Results}

\subsection{ECG FMs have distinct depth--scale decomposition surfaces}

Figure~\ref{fig:semantic-atlas} summarizes single-coordinate clinical accessibility at three resolutions: test depth--scale surfaces for every FM, paired patient-bootstrap common-scale semantic AUCs, and concept-family profiles. The corresponding reconstruction and dead-feature atlases are included in Appendix~\ref{app:additional-atlases}. These surfaces are the primary result object: a model can localize clinical measurements more strongly at one depth while requiring a wider dictionary or exhibiting less stable atom identity elsewhere. We therefore retain all 25 depth--scale cells per model as the prespecified multiscale evaluation surface.

Interpretability is consequently represented as a structured profile of sparse reconstructability, single-coordinate clinical accessibility, coverage, and feature reproducibility. Common-scale AUC provides a capacity-robust comparison within each axis, while the underlying depth--scale surfaces support operating-point selection for specific analysis goals.

\subsection{Matched-scale curves produce metric-specific FM orderings}

Figure~\ref{fig:matched-curves} compares all FMs only at the same $E$. Shaded regions are paired patient-cluster intervals after averaging the three fixed SAE seeds and five target depths. For reconstruction, the highest common-scale point estimate is \MSReconTopModel{} (AUC \MSReconTopAUC{}, 95\% CI [\MSReconTopCILow{}, \MSReconTopCIHigh{}], bootstrap rank-1 probability \MSReconTopRankOne{}). The complete reconstruction order is \MSReconOrder{}, with \MSReconSignificantPairs{} of 15 pairwise contrasts surviving BH correction.

Single-coordinate clinical accessibility has a separate ordering: \MSSemanticOrder{}, with \MSSemanticSignificantPairs{} of 15 FM contrasts surviving BH correction. The matched model-level profile is complemented by substantial concept-level resolution: for \MSSemanticTopModel{}, the 49 concept-specific common-scale AUCs have median 0.406 and interquartile range 0.250--0.442; 26 concepts reach at least 0.40, eight reach at least 0.50, and four reach at least 0.60. Ventricular rate, QRS duration, P-wave detection, and mean RR are among the most locally accessible concepts. Concept coverage similarly orders the models as \MSCoverageOrder{}, with \MSCoverageSignificantPairs{} BH-significant pairwise contrasts. Exact common-scale AUC estimates and paired confidence intervals are provided in Appendix Table~\ref{tab:patient-profiles}.

Appendix~\ref{app:mimic-replication} reports the MIMIC-IV-ECG replication, preserving metric-specific leaders. Table~\ref{tab:input-harmonization} summarizes the final-layer input-harmonization audit detailed in Appendix~\ref{app:input-harmonization}; the six-model accessibility order is preserved under the harmonized protocols.

\begin{figure*}[t]
  \centering
  \includegraphics[width=0.86\textwidth]{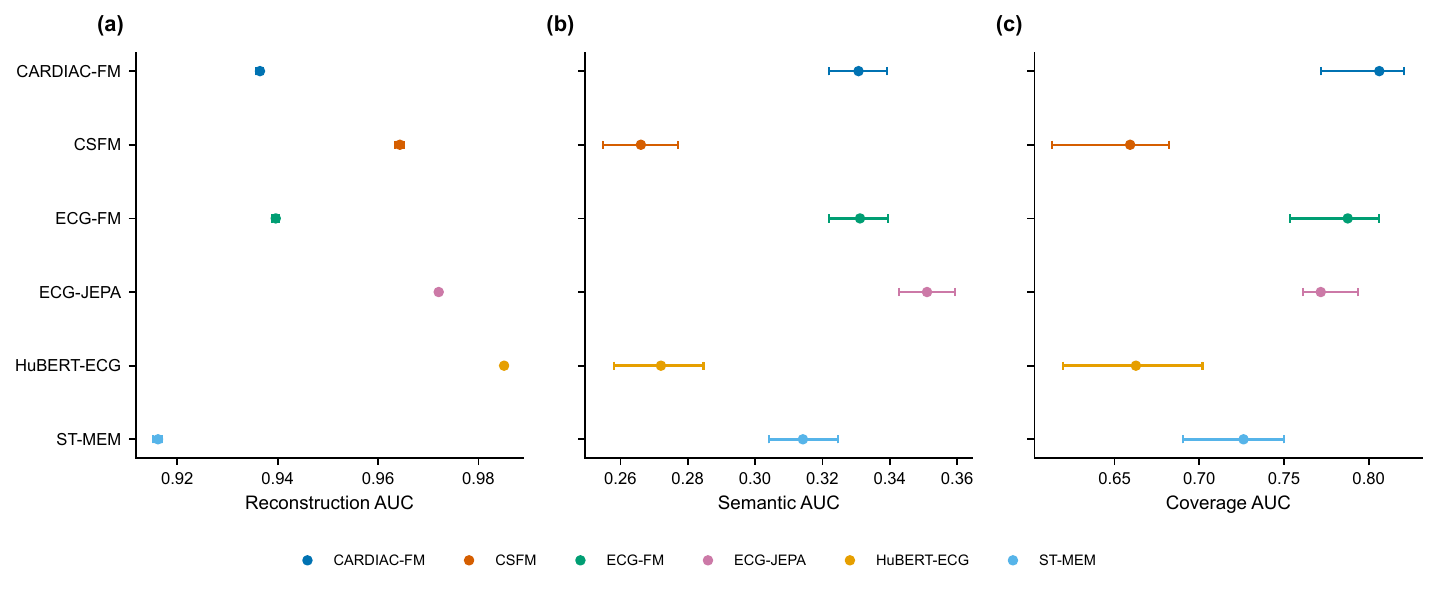}
  \caption{Metric-specific common-scale AUC estimates and paired patient-cluster intervals for (a) reconstruction, (b) single-coordinate clinical accessibility, and (c) concept coverage. A higher value is favorable within each panel, and the panels jointly characterize complementary dimensions of representation interpretability.}
  \Description{Three forest plots compare six ECG foundation models on reconstruction, single-coordinate clinical accessibility, and concept coverage after integrating the same five SAE scales.}
  \label{fig:profile-forest}
\end{figure*}

\subsection{External validation preserves metric-specific leaders}

We repeated the complete 450-cell multiscale SAE audit on a credentialed, patient-partitioned MIMIC-IV-ECG cohort; cohort construction and sensitivity analyses are detailed in Appendix~\ref{app:mimic-replication}. Table~\ref{tab:mimic-model-profiles} reports the held-out model profiles after integrating the same five-point expansion curve and averaging over the same five relative depths.

The replication reveals complementary leaders across interpretability dimensions. HuBERT-ECG has the highest reconstruction profile (0.990), whereas ECG-JEPA has the highest single-feature semantic alignment (0.324) and concept coverage (0.781). ECG-FM has the lowest dead-feature fraction (0.099) and the highest feature stability (0.378), while CSFM has the highest subspace stability (0.778). Together, reconstruction fidelity, local clinical accessibility, dictionary utilization, and cross-seed reproducibility provide a multidimensional comparison of the six models.

\subsection{A calibration ladder separates availability from localization}

Detailed calibration results are provided in Appendix Table~\ref{tab:accessibility-calibration}. The full SAE retains 79.9--87.4\% of dense-ridge mean $|r|$. Single SAE coordinates outperform the mean of 20 matched random dictionaries by 0.034--0.088 across every model's five depths, with 78.8--87.8\% concept--depth wins. These held-out results demonstrate consistent recovery of clinically aligned sparse structure by the learned SAE dictionaries across all six ECG FMs.

This calibration connects distributed concept availability with sparse localization. Full-code readouts quantify preserved clinical information, and single-coordinate readouts identify aligned sparse features across all six FMs. Together, these readouts establish the SAE profile as a common sparse instrument for comparing FMs.

The single-coordinate profile quantifies how strongly each clinical measurement is localized to one train-selected SAE coordinate and retained under held-out evaluation. Its 49-concept macro-average supports standardized model-level comparison, while the concept-level distribution reveals a substantial high-accessibility subset and resolves differences across waveform measurements. Together, these views provide both a common summary and concept-specific resolution of sparse clinical accessibility.

\subsection{Rank stability is itself a benchmark result}

Rank stability across SAE capacities is quantified over all ten pairs of common $E$ values. Reconstruction rank agreement has median Kendall $\tau=\MSReconMedianScaleTau$ and minimum $\tau=\MSReconMinScaleTau$; single-coordinate accessibility has median $\tau=\MSSemanticMedianScaleTau$ and minimum $\tau=\MSSemanticMinScaleTau$; coverage has median $\tau=\MSCoverageMedianScaleTau$ and minimum $\tau=\MSCoverageMinScaleTau$. Figure~\ref{fig:profile-forest} presents common-scale AUC as a capacity-robust summary, complemented by full curves for decisions tied to a particular SAE budget.

Rank agreement across $E$ directly tests whether an FM conclusion survives measurement capacity. The shared expansion grid provides a capacity-aware assessment of FM ordering under a common evaluation protocol. The current atlas also matches absolute width because every audited layer has $d=768$.

\begin{figure*}[t]
  \centering
  \includegraphics[width=\textwidth]{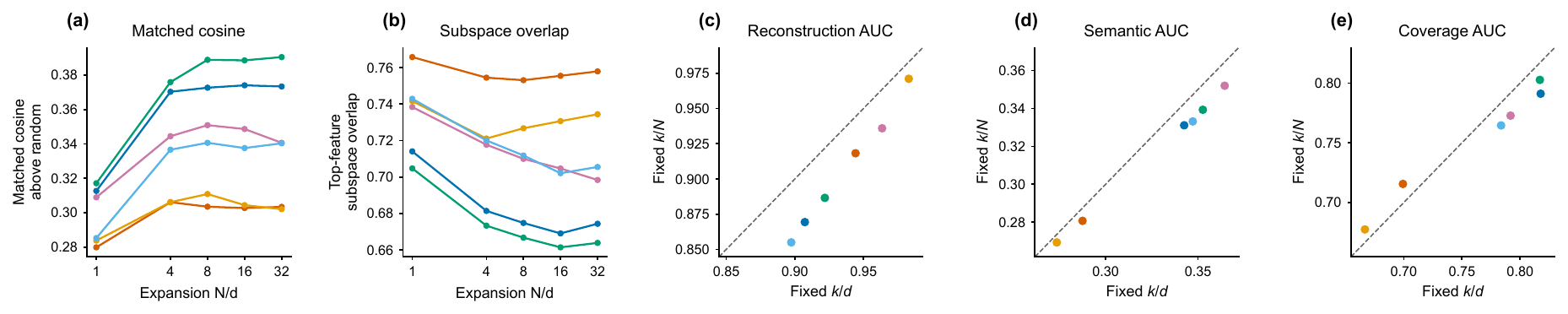}
  \includegraphics[width=0.74\textwidth]{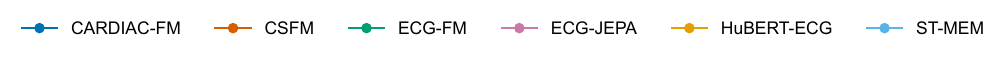}
  \caption{Robustness axes. (a) Cross-seed decoder matching above random and (b) selected-subspace overlap over common scale. Panels (c)--(e) show mid-depth common-scale AUC for reconstruction, clinical accessibility, and concept coverage under fixed $k/d$ and fixed $k/N$; the diagonal denotes unchanged profiles.}
  \Description{Five equal-size panels show feature stability curves and sparsity-sensitivity scatter plots for six models, with a separate legend below the panels.}
  \label{fig:stability-sensitivity}
\end{figure*}

\subsection{Feature identity and sparse fidelity are different axes}

Cross-seed matching asks whether a favorable fidelity or accessibility profile is implemented by reproducible decoder directions. The highest above-random matched-cosine AUC belongs to \MSStabilityTopModel{} (\MSStabilityTopAUC{}), whereas the highest selected-subspace overlap AUC belongs to \MSSubspaceTopModel{} (\MSSubspaceTopAUC{}). Figure~\ref{fig:stability-sensitivity}(a)--(b) shows how both quantities change over common scale. Joint reporting of stability, reconstruction, and single-coordinate clinical accessibility captures complementary representation properties.

Feature-level matching and selected-subspace overlap resolve two complementary forms of reproducibility across SAE seeds.

\subsection{Model ordering is tested under a second sparsity parameterization}

To isolate the effect of the active-code budget, we compare the primary fixed-$k/d$ arm with a preregistered fixed-$k/N$ arm at relative depth $q=0.5$. Because $d=768$, the primary arm uses $k=96$ at every $E$, whereas the sensitivity arm uses $k=\{12,48,96,192,384\}$ for $E=\{1,4,8,16,32\}$; dictionary widths, models, seeds, and training settings remain matched. Reconstruction ordering is highly stable across the two arms (median/minimum Kendall $\tau=1.00/0.87$), while accessibility ($0.73/0.60$) and coverage ($0.87/0.47$) respond more to the active budget, with the largest coverage reordering at $E=16$. Relative to fixed-$k/d$, fixed-$k/N$ changes common-scale AUC by $-0.012$ to $-0.042$ for reconstruction, $-0.004$ to $-0.014$ for accessibility, and $-0.027$ to $+0.016$ for coverage. At the shared $E=8$ anchor, both arms use $N=6144,k=96$, and independently trained cells differ by at most $0.0272$. Figure~\ref{fig:stability-sensitivity}(c)--(e) displays the corresponding model-level profiles.

\section{Limitations and Claim Boundary}

\textbf{First,} the current benchmark covers six representative ECG foundation models spanning a diverse set of encoder architectures and pretraining objectives. Although this collection enables controlled comparisons across these models, future versions could incorporate newly released ECG foundation models and additional model scales to further broaden the benchmark.

\textbf{Second,} the clinical concept inventory emphasizes waveform measurements and diagnostic concepts that can be evaluated consistently across models and cohorts. Extending the inventory to finer-grained morphological patterns, continuous physiological attributes, longitudinal changes, and expert-annotated concepts could provide a more comprehensive view of representation interpretability.

\textbf{Third,} ECG-InterpBench uses a standardized SAE capacity and sparsity grid to ensure matched comparisons across encoders. Future work could evaluate finer-grained capacity ranges, alternative sparsity objectives, and additional dictionary-learning methods while preserving the same capacity-controlled evaluation protocol. These extensions would also facilitate adaptation of the benchmark to foundation models for EEG, medical imaging, electronic health records, and other clinical modalities.

Throughout, ECG-InterpBench characterizes the structure of frozen model representations under a matched SAE protocol; it does not establish physiological mechanisms, clinical utility, or prospective patient outcomes. The reported model orderings are therefore specific to the evaluated representation properties and should not be interpreted as rankings of diagnostic performance or overall clinical value.

\FloatBarrier

\section{Conclusion}

\bench{} uses a calibrated multi-scale SAE family to audit and compare ECG foundation-model representations. Its core unit is an exact six-model matched block. The resulting depth--scale profiles, paired patient inference, seed stability, sparsity sensitivity, and dense-to-single calibration distinguish faithful sparse reconstruction, distributed concept availability, single-coordinate accessibility, and reproducible feature identity. This design turns ECG-FM interpretability into a systematic and reproducible model-comparison problem. More broadly, \bench{} provides a general framework for identifying where, at what capacity, and how consistently clinically meaningful structure emerges within foundation-model representations. By connecting representation geometry with clinical concepts across models and scales, it establishes a foundation for the principled development and selection of interpretable ECG foundation models. The same capacity-controlled auditing framework can be extended to other clinical foundation models, including those for medical imaging, electroencephalography, electronic health records, physiological time series, and multimodal patient data. With modality-specific clinical concepts and matched representation interfaces, this framework can support systematic interpretability comparisons across a broader range of medical AI systems.

\section{Ethics and Responsible Use}

All analyses use de-identified secondary ECG datasets under their respective licenses and data-access agreements. No new participants were recruited, and no clinical intervention was performed. PTB-XL provides the directly accessible public reproduction path for the benchmark, while MIMIC-IV-ECG is a publicly available dataset that requires credentialed access through an application and approval process. Released artifacts exclude restricted waveforms, protected metadata, and record-level MIMIC identifiers.

ECG-InterpBench is a research benchmark for evaluating the representation-level interpretability of ECG foundation models. Its outputs are not intended for direct use in diagnosis, treatment, patient management, or other clinical decision-making settings.

The accompanying artifact includes executable evaluation code, standardized manifests, aggregate and configuration-level metrics, fixed random and bootstrap seeds, audit outputs, and figure-generation scripts. The public PTB-XL workflow supports end-to-end reproducible evaluation, while the MIMIC-IV-ECG analysis provides an additional reproduction path for approved credentialed users. Model weights and source datasets remain governed by their original repositories, licenses, and access requirements.

\clearpage
\bibliographystyle{ACM-Reference-Format}
\bibliography{references}

\clearpage
\appendix
\setcounter{figure}{0}
\renewcommand{\thefigure}{A\arabic{figure}}
\providecommand{\theHfigure}{}
\renewcommand{\theHfigure}{A\arabic{figure}}
\setcounter{table}{0}
\renewcommand{\thetable}{S\arabic{table}}
\providecommand{\theHtable}{}
\renewcommand{\theHtable}{S\arabic{table}}

\section{Common-Scale Model Profiles}
\label{app:patient-profiles}

Table~\ref{tab:patient-profiles} reports the exact model-level common-scale AUC estimates and paired patient-cluster confidence intervals underlying the visual comparison in Figure~\ref{fig:profile-forest}.

\begin{table*}[!t]
  \centering
  \small
  \caption{Common-five-scale model profiles with paired patient-cluster 95\% intervals. The semantic column reports the train-selected single-coordinate $|r|$ AUC defined in Section~\ref{sec:benchmark-design}. Each AUC integrates the same $E\in\{1,4,8,16,32\}$ support and averages the same five depth targets and three SAE seeds.}
  \label{tab:patient-profiles}
  \begin{tabular}{lccc}
\toprule
Model & Recon AUC & Semantic AUC & Coverage AUC \\
\midrule
CARDIAC-FM & 0.936 [0.936, 0.937] & 0.331 [0.322, 0.339] & \textbf{0.806} [0.772, 0.821] \\
CSFMe & 0.964 [0.963, 0.965] & 0.266 [0.255, 0.277] & 0.659 [0.613, 0.682] \\
ECG-FM & 0.940 [0.939, 0.940] & 0.331 [0.322, 0.340] & 0.788 [0.753, 0.806] \\
ECG-JEPA & 0.972 [0.972, 0.972] & \textbf{0.351} [0.343, 0.359] & 0.772 [0.761, 0.794] \\
HuBERT-ECG & \textbf{0.985} [0.985, 0.985] & 0.272 [0.258, 0.285] & 0.663 [0.620, 0.702] \\
ST-MEM & 0.916 [0.915, 0.917] & 0.314 [0.304, 0.325] & 0.726 [0.690, 0.750] \\
\bottomrule
\end{tabular}
{\centering Audited patient-profile table pending.}}%
  }
\end{table*}

\FloatBarrier
\section{Fixed-Scale Accessibility Calibration}
\label{app:accessibility-calibration}

Table~\ref{tab:accessibility-calibration} reports the complete held-out calibration that compares distributed ridge readouts with increasingly localized SAE, native-coordinate, and random-coordinate readouts at the common $E=8$ scale.

\begin{table*}[!b]
  \centering
  \small
  \setlength{\tabcolsep}{3pt}
  \caption{Held-out $E=8$ accessibility calibration over five depths and 49 concepts. Ridge columns report mean $|r|$. One-dimensional entries report mean $|r|$/coverage at $|r|\geq.20$; SAE values average three seeds and random values average 20 dictionaries. All coordinates are selected on training data only.}
  \label{tab:accessibility-calibration}
  \begin{tabular}{lrrrrrrr}
\toprule
Model & Dense ridge & Dense 1D & Full SAE & Top-16 & Top-4 & SAE 1D & Random 1D \\
\midrule
CARDIAC-FM & 0.764 & 0.377/0.927 & \textbf{0.677} & 0.514 & 0.423 & 0.320/\textbf{0.796} & 0.232/0.591 \\
CSFM & 0.651 & 0.301/0.739 & 0.536 & 0.410 & 0.341 & 0.266/0.661 & 0.190/0.480 \\
ECG-FM & \textbf{0.766} & 0.385/\textbf{0.943} & 0.675 & 0.513 & 0.417 & 0.321/0.778 & 0.243/0.634 \\
ECG-JEPA & 0.726 & \textbf{0.421}/0.910 & 0.645 & \textbf{0.535} & \textbf{0.430} & \textbf{0.342}/0.758 & \textbf{0.290}/\textbf{0.697} \\
HuBERT-ECG & 0.665 & 0.325/0.776 & 0.565 & 0.412 & 0.340 & 0.265/0.654 & 0.231/0.562 \\
ST-MEM & 0.725 & 0.379/0.886 & 0.615 & 0.474 & 0.391 & 0.311/0.714 & 0.245/0.572 \\
\bottomrule
\end{tabular}
{\centering Audited accessibility-calibration table pending.}}%
  }
\end{table*}

\clearpage
\section{Frozen Source-Layer Interfaces}
\label{app:model-interfaces}

Table~\ref{tab:multiscale-models} defines the representation-extraction contract used throughout the benchmark. Each released checkpoint is frozen and evaluated through its native lead selection, temporal resolution, and tokenization path. The models consequently receive inputs ranging from eight to twelve leads and from 2,250 to 5,000 samples, while the audited states are taken before any model-specific downstream projection that would change the source width. Preserving these interfaces measures the representations available under realistic model use rather than introducing an untrained common tokenizer.

\begin{table*}[!t]
\centering
\small
\caption{Frozen source-layer interfaces. The benchmark preserves each released ECG input path while matching the analyzed layer width. ``Layers'' counts available audited states, including the indexed boundary states used by the released extraction code.}
\label{tab:multiscale-models}
\begin{tabular}{@{}lp{3.0cm}p{3.0cm}p{4.2cm}cc@{}}
\toprule
\textbf{Model} & \textbf{Pretraining family} & \textbf{Input used} & \textbf{Audited layer state} & \textbf{Layers} & $\boldsymbol{d}$ \\
\midrule
CSFM & multimodal sensing & $12\times2500$ & released encoder/classification stream & 6 & \textbf{768} \\
CARDIAC-FM & ECG+MRI risk & $12\times5000$ & transformer state before 768-to-512 pool projection & \textbf{12} & \textbf{768} \\
ECG-FM & wav2vec and contrastive & $12\times5000$ & token states & \textbf{12} & \textbf{768} \\
ECG-JEPA & joint-embedding prediction & $8\times2500$ & encoder states after released lead selection & \textbf{12} & \textbf{768} \\
HuBERT-ECG & masked clustering & flattened $12\times2500$ & token states & \textbf{12} & \textbf{768} \\
ST-MEM & spatio-temporal masking & $12\times2250$ & \texttt{forward\_encoding} states & \textbf{12} & \textbf{768} \\
\bottomrule
\end{tabular}
\end{table*}

The shared audited width $d=768$ permits exact capacity matching: the five expansion factors correspond to absolute dictionary widths $N\in\{768,3072,6144,12288,24576\}$, with the primary active budget fixed at $k=96$. Relative depth maps the six available CSFM states and the twelve states exposed by each other encoder onto the same five index positions, providing a common coordinate over each encoder trajectory. Table~\ref{tab:multiscale-models} specifies the resulting representation interfaces, and Appendix~\ref{app:input-harmonization} tests how heterogeneous waveform interfaces affect the final-layer model ordering.

\FloatBarrier
\section{Additional Depth--Scale Atlases}
\label{app:additional-atlases}

\begin{figure*}[!t]
  \centering
  \includegraphics[width=\textwidth]{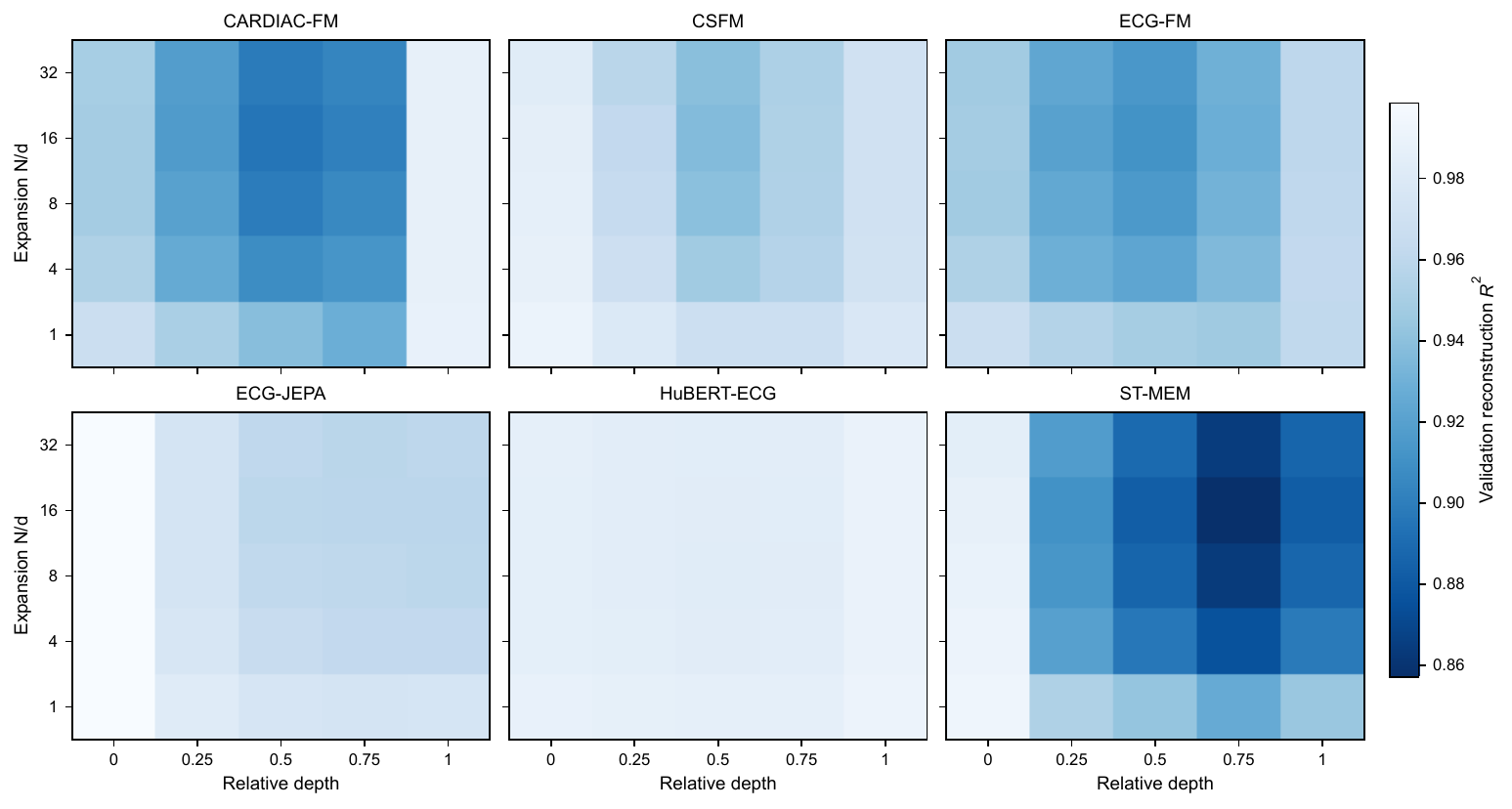}
  \caption{Validation reconstruction $R^2$ over the complete matched depth--scale atlas.}
  \label{fig:reconstruction-atlas}
  \Description{Six heatmaps show sparse-autoencoder reconstruction over five relative depths and five exactly matched expansion factors.}
\end{figure*}

\begin{figure*}[!t]
  \centering
  \includegraphics[width=\textwidth]{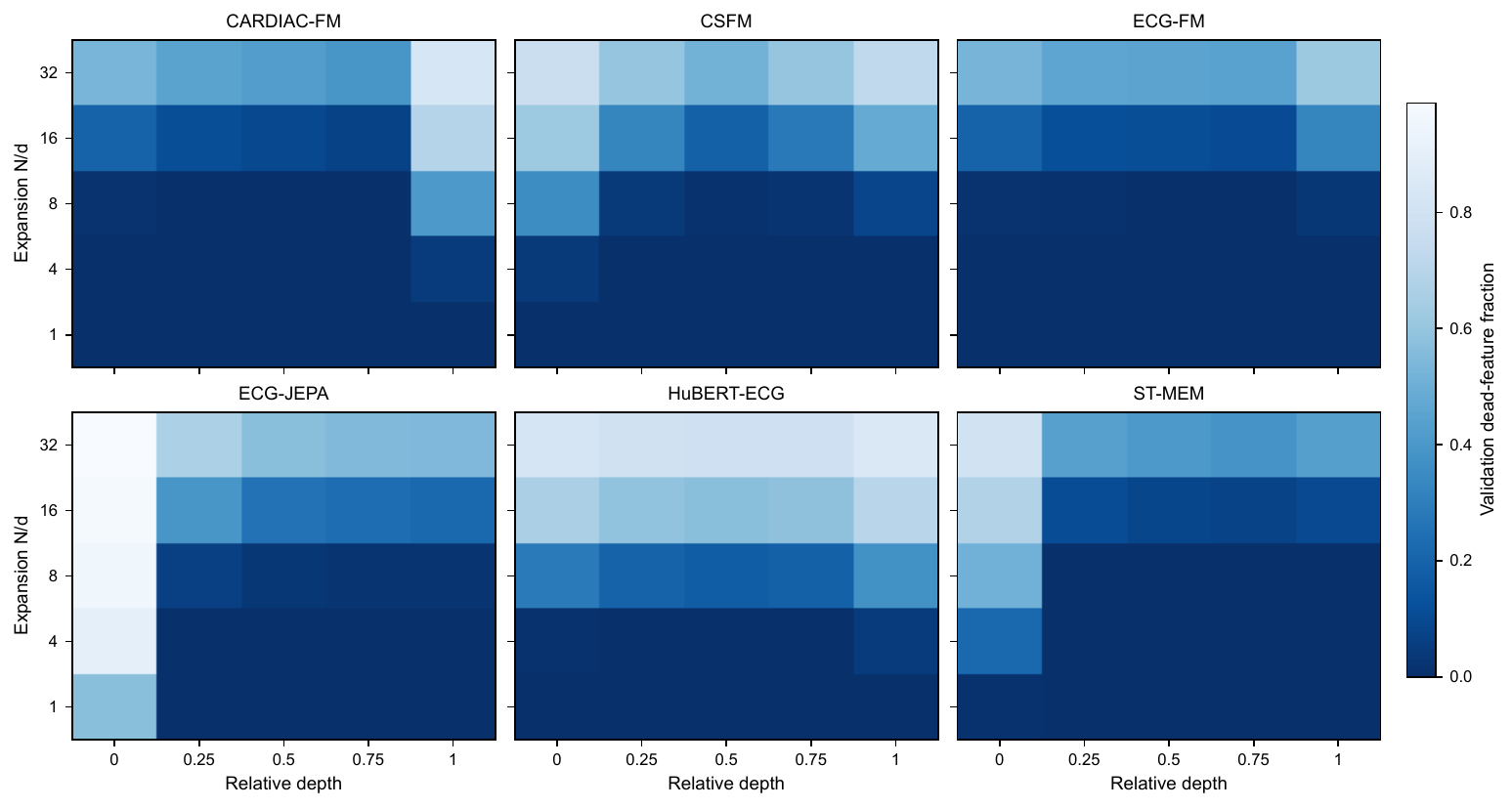}
  \caption{Validation dead-feature fraction over the complete matched depth--scale atlas.}
  \label{fig:dead-feature-atlas}
  \Description{Six heatmaps show dead-feature fractions over five relative depths and five exactly matched expansion factors.}
\end{figure*}

Figures~\ref{fig:reconstruction-atlas} and~\ref{fig:dead-feature-atlas} expose the two validation fidelity diagnostics for all 150 model--depth--scale configurations underlying the 450 trained SAE cells. Each heatmap cell averages the three fixed SAE seeds, and each metric uses one shared color scale across all six models. These complete surfaces complement the clinical-accessibility atlas in Figure~\ref{fig:semantic-atlas} and prevent a favorable depth or dictionary width from being selected in isolation.

Reconstruction remains high overall but does not improve monotonically with dictionary width under the fixed $k=96$ budget. Averaged descriptively over all models, depths, and seeds, validation $R^2$ is 0.968, 0.953, 0.949, 0.947, and 0.948 for $E=1,4,8,16,$ and 32, respectively. The model-averaged profiles range from 0.918 for ST-MEM to 0.986 for HuBERT-ECG, while the depth average falls from 0.978 at the initial state to approximately 0.937 at relative depths 0.5--0.75 before reaching 0.962 at the final state. Thus additional dictionary coordinates do not substitute for model- and depth-specific reconstruction structure when the active-code budget is held constant.

Dictionary utilization changes much more strongly with scale. The corresponding mean dead-feature fractions are 0.019, 0.042, 0.127, 0.338, and 0.604 from $E=1$ through $E=32$. Model-level averages range from 0.137 for ECG-FM to 0.337 for HuBERT-ECG, and the initial-state average (0.388) is substantially higher than the averages near relative depths 0.5 and 0.75 (0.156 and 0.158). HuBERT-ECG therefore combines the highest descriptive reconstruction average with the highest dead-feature average, directly illustrating that sparse fidelity and dictionary utilization are distinct properties. Reconstruction, dead-feature fraction, and the preregistered validation fidelity gate jointly characterize operating points across the complete dictionary-width curve.

\FloatBarrier
\section{Credentialed MIMIC-IV-ECG Replication}
\label{app:mimic-replication}

We repeated the multiscale SAE audit on the credentialed MIMIC-IV-ECG cohort~\cite{gow2023mimic}. The fixed manifest contains 100,000 ECGs from 48,491 patients, partitioned by patient into 70,260 training, 9,768 validation, and 19,972 test records; 54,591 records have complete waveform-derived measurements. Source waveforms and identifiers remain in the credentialed environment. The replication uses the same six frozen encoders, relative depths $\{0,.25,.5,.75,1\}$, expansion factors $E\in\{1,4,8,16,32\}$, $k=96$, three SAE seeds, and 8,000 training steps as the primary benchmark, for 450 completed SAE cells. For this SAE-only analysis, semantic alignment and coverage use seven continuous waveform concepts: heart rate, QRS duration, PR interval, a QT-like interval, and global R-, ST-, and T-amplitude measurements.

The model-level held-out profiles are reported in Table~\ref{tab:mimic-model-profiles}; this appendix provides the cohort, uncertainty, and scale--depth sensitivity details.

Patient-cluster uncertainty uses 2,000 test-set bootstrap draws over 6,609 patients, retaining all ECGs from each sampled patient. The leading profiles are precise: HuBERT-ECG reconstruction is 0.9905 (95\% CI 0.9904--0.9906), while ECG-JEPA semantic alignment is 0.3244 (0.3190--0.3299) and coverage is 0.7810 (0.7400--0.8057). The intervals quantify patient-sampling uncertainty conditional on the frozen encoders, trained SAEs, and training-selected coordinates.

The aggregate sensitivity analysis in Table~\ref{tab:mimic-scale-depth} demonstrates the resolution gained from a multiscale, multidepth audit. Across expansion factors, reconstruction remains near 0.97, semantic alignment is highest at $E=1$, and the dead-feature fraction traces dictionary utilization from 0.032 at $E=1$ to 0.636 at $E=32$. Across depth, semantic alignment is highest near relative depth 0.5 and coverage peaks near 0.75. These profiles show how MIMIC representation properties vary jointly with SAE capacity and encoder depth.

\begin{table*}[t]
  \centering
  \small
  \caption{MIMIC-IV-ECG scale and depth sensitivity, aggregated across the six models. Expansion rows average over five depths; depth rows average over five expansions. All rows average the three SAE seeds.}
  \label{tab:mimic-scale-depth}
  \begin{tabular}{llrrrr}
    \toprule
    Axis & Setting & Recon. $R^2$ & Semantic alignment & Coverage & Dead fraction \\
    \midrule
    Expansion & $E=1$  & 0.973 & \textbf{0.299} & \textbf{0.710} & 0.032 \\
              & $E=4$  & 0.971 & 0.281 & 0.624 & 0.134 \\
              & $E=8$  & 0.971 & 0.284 & 0.632 & 0.277 \\
              & $E=16$ & 0.971 & 0.284 & 0.637 & 0.461 \\
              & $E=32$ & 0.970 & 0.283 & 0.627 & \textbf{0.636} \\
    \midrule
    Relative depth & 0    & \textbf{0.986} & 0.295 & 0.640 & 0.542 \\
                   & 0.25 & 0.973 & 0.297 & 0.662 & 0.293 \\
                   & 0.50 & 0.964 & 0.298 & 0.689 & 0.211 \\
                   & 0.75 & 0.959 & 0.289 & 0.692 & 0.194 \\
                   & 1.00 & 0.973 & 0.252 & 0.546 & 0.301 \\
    \bottomrule
  \end{tabular}
\end{table*}

Matched decoder features vary across seeds (model-level stability AUC 0.293--0.378), whereas their selected subspaces are more reproducible (overlap AUC 0.702--0.778). This distinction identifies the selected subspace as the more stable unit for cross-seed interpretation and supports model-level conclusions about reproducible representation structure. The release audit passes all requirements with 450/450 SAE cells and no missing matched depth--scale block. Together, these results provide a complete credentialed-cohort replication of the benchmark profile.

\section{Input-Interface Harmonization Audit}
\label{app:input-harmonization}

The primary benchmark preserves each released model's input interface to reflect realistic use, but this choice could confound encoder differences with lead selection, sampling, or windowing. We therefore repeated final-layer dense-coordinate accessibility under four protocols on the same 21,799 PTB-XL records and patient split. \emph{Native} retains each released interface. \emph{Lead} supplies the common independent leads I, II, and V1--V6 and reconstructs the dependent limb leads for 12-lead models using
$\mathrm{III}=\mathrm{II}-\mathrm{I}$,
$\mathrm{aVR}=-(\mathrm{I}+\mathrm{II})/2$,
$\mathrm{aVL}=\mathrm{I}-\mathrm{II}/2$, and
$\mathrm{aVF}=\mathrm{II}-\mathrm{I}/2$.
\emph{Temporal} maps the complete 10-second waveform to a common 250-Hz, 2,500-point grid before adapting its length to each checkpoint. \emph{Joint} applies both transformations. All harmonized waveforms are constructed in physical units and standardized per lead. Tokenizers, patch embeddings, and pretrained weights remain model-native; this is therefore a waveform-harmonized, tokenizer-native audit rather than a shared-tokenizer experiment.

For each of 49 waveform concepts, the best dense coordinate and its sign are selected only on a fixed 4,096-record subset of the training split. Mean absolute correlation and coverage at $|r|\geq0.20$ are then evaluated without reselection on 3,325 patient-disjoint test records. We additionally compare each harmonized representation with its native counterpart using linear CKA. Uncertainty uses 2,000 patient-cluster bootstrap draws, with Benjamini--Hochberg correction applied across the 18 model-level comparisons and, separately, across all concept-level comparisons.

The model-level results are reported in Table~\ref{tab:input-harmonization}.

\begin{figure*}[t]
  \centering
  \includegraphics[width=\textwidth]{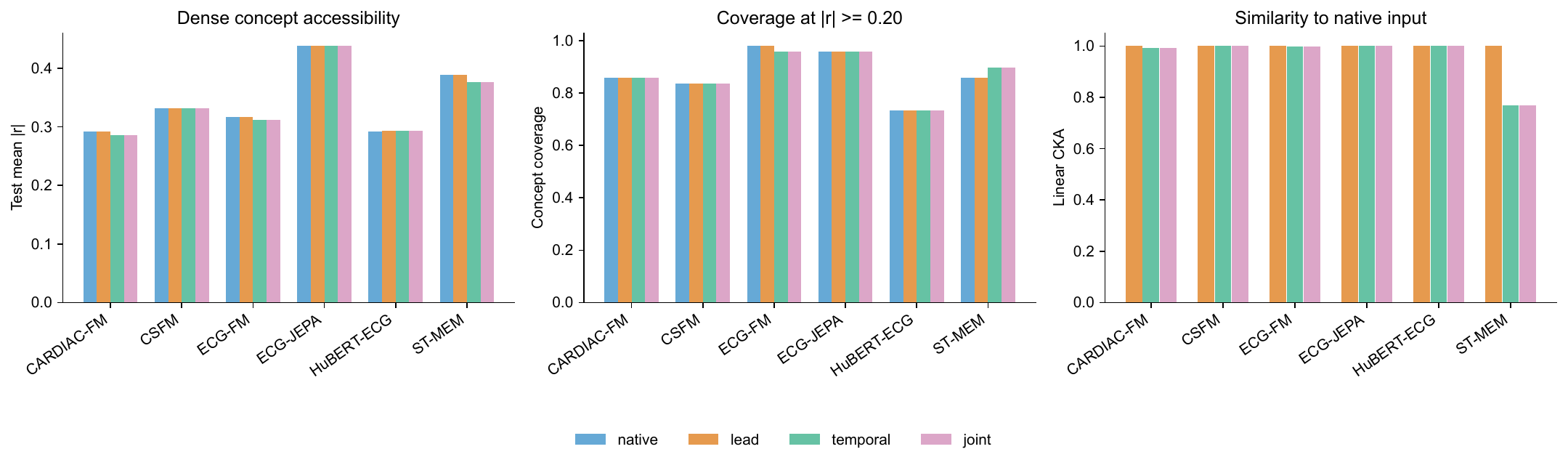}
  \caption{Final-layer accessibility under native, lead-harmonized, temporal-harmonized, and jointly harmonized inputs. Left: test mean absolute correlation after training-only coordinate selection. Center: concept coverage at $|r|\geq0.20$. Right: linear CKA between each harmonized representation and its native counterpart.}
  \Description{Three grouped bar charts compare six ECG foundation models under four input protocols. Lead harmonization leaves accessibility and representations nearly unchanged. Temporal and joint harmonization modestly lower accessibility for CARDIAC-FM and ECG-FM and produce the largest representation change for ST-MEM.}
  \label{fig:input-harmonization}
\end{figure*}

Lead harmonization preserves mean $|r|$ within 0.0006 and achieves lead-to-native CKA of at least 0.9999. Temporal harmonization yields changes of $\Delta=-0.0058$ for CARDIAC-FM (95\% CI $[-0.0095,-0.0022]$, $q=0.016$), $\Delta=-0.0048$ for ECG-FM (95\% CI $[-0.0083,-0.0013]$, $q=0.016$), and $\Delta=-0.0131$ for ST-MEM (95\% CI $[-0.0196,-0.0067]$, $q=0.009$). The nearly identical joint and temporal effects identify time-grid and window handling as the source of the measurable changes. ST-MEM's joint CKA of 0.770 reflects the contrast between its native centered 9-second crop and the harmonized complete 10-second record mapped to 2,250 input points.

Model-level coverage is stable after FDR correction (all $q=1.0$), and the six-model ordering by mean accessibility is identical under every protocol (Spearman $\rho=1.0$; Kendall $\tau=1.0$). Coverage ordering is also stable (Spearman $\rho=0.971$ for temporal and joint). ECG-JEPA and HuBERT-ECG mean differences range from $3.5\times10^{-5}$ to $5.7\times10^{-4}$, further demonstrating the close agreement between native and harmonized accessibility profiles.

Overall, the stable ordering across all four input protocols indicates that the observed cross-model differences primarily reflect encoder-representation structure. Lead harmonization has negligible effect, and temporal harmonization preserves the ordering across models. This audit strengthens the attribution of the benchmark profiles to the learned representations.

\section{Validation-Matched SAE--Dense Intervention Selectivity}
\label{app:matched-effect}

The accessibility analyses ask whether a concept can be read from one coordinate. We additionally test whether a sparse coordinate can change a frozen concept readout with less spillover than a native dense coordinate. This is a final-layer, matched-effect comparison over the same 49 waveform concepts. Both methods receive 768 candidate coordinates: the native FM dimensions for Dense and a fixed, target-independent 768-coordinate subset of the $E=8$ dictionary for SAE. For each concept, the top five coordinates and the high-concept centroid are selected using training data only. Intervention doses are then calibrated on validation data to a common achievable target-readout change, capped at $+0.25$ standard deviations, with a minimum feasible change of $0.05$ and interpolation coefficient at most one. Features, centroids, and doses are frozen before patient-disjoint test evaluation.

The SAE arm uses three training seeds and 20 fixed candidate subsets per seed. Primary inference requires a concept to pass the validation effect gate for all three seeds; failed concepts remain in the fixed denominator and are not replaced. The two primary endpoints are cross-family off-target root mean square change and the corresponding wrong-behavior index (WBI), defined as off-target RMS divided by the absolute target effect. Lower values indicate a more selective intervention. Confidence intervals and two-sided tests use 2,000 paired patient-cluster bootstrap draws, followed by Benjamini--Hochberg correction over the preregistered metric families.

\begin{table*}[t]
  \centering
  \caption{Final-layer matched-effect comparison at $k=5$ across the five models satisfying the all-seed validation gate. The eligibility count $n$ requires all three SAE seeds to pass the gate. WBI and deltas are measured on the patient-disjoint test split; negative SAE--Dense deltas favor SAE.}
  \label{tab:matched-effect-dense-sae}
  \begin{tabular}{lrrrrr}
    \toprule
    Model & $n$ & Dense WBI & SAE WBI & $\Delta$ off-target RMS & $\Delta$ WBI \\
    \midrule
    CARDIAC-FM & 5  & 1.614 & 0.971 & $-0.046$ & $-0.643$ \\
    CSFM       & 11 & \textbf{2.134} & 0.944 & $-0.087$ & $-1.189$ \\
    ECG-JEPA   & \textbf{15} & 1.436 & 0.691 & $-0.057$ & $-0.746$ \\
    HuBERT-ECG & 10 & 1.315 & \textbf{1.138} & $\boldsymbol{-0.014}$ & $\boldsymbol{-0.176}$ \\
    ST-MEM     & 10 & 1.678 & 0.736 & $-0.063$ & $-0.942$ \\
    \bottomrule
  \end{tabular}
\end{table*}

\begin{figure*}[t]
  \centering
  \includegraphics[width=0.9\textwidth]{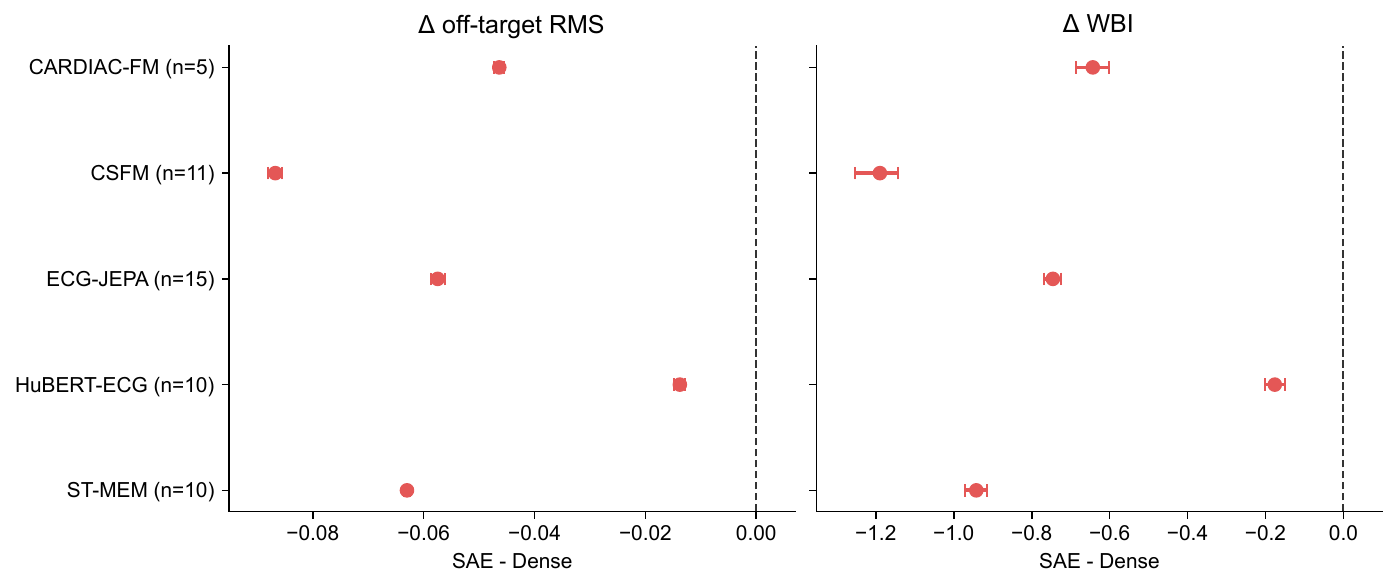}
  \caption{SAE--Dense differences in cross-family off-target RMS and WBI with paired patient-cluster 95\% confidence intervals. Negative values favor SAE; model labels give the number of concepts satisfying the strict all-seed gate.}
  \Description{Two forest plots show negative SAE-minus-Dense differences and paired 95 percent confidence intervals for off-target RMS and wrong-behavior index across five models with validation-qualified concepts.}
  \label{fig:matched-effect-sae-dense-deltas}
\end{figure*}

Figure~\ref{fig:matched-effect-sae-dense-deltas} visualizes the paired bootstrap contrasts. All ten 95\% confidence intervals lie entirely below zero, showing consistent SAE reductions in both off-target RMS and WBI across the five models. Off-target RMS differences range from $-0.014$ for HuBERT-ECG to $-0.087$ for CSFM, while WBI differences range from $-0.176$ to $-1.189$ for the same models. CSFM shows the largest reduction on both endpoints, and the favorable direction remains consistent across eligibility sets ranging from 5 to 15 concepts. Across the resulting model--concept evaluations, SAE has a favorable and FDR-significant difference in all 10 model--metric tests ($q\leq .0011$). At a validation-matched frozen readout effect, the selected SAE coordinates consistently produce less cross-family readout spillover than native dense coordinates.

\end{document}